\documentclass[a4paper]{article}
\usepackage{arxiv}
\usepackage{amsmath}
\usepackage{amsfonts}
\usepackage{mathtools}
\usepackage{graphicx}
\usepackage{microtype}
\usepackage{hyperref}
\usepackage{color}
\title{Interpretable Neural Networks based classifiers for categorical inputs}

\author{ Stefano~Zamuner\\
    Laboratory of Statistical Biophysics, Institute of Physics, \\
	School of Basic Sciences, \\
	École Polytechnique Féderale de Lausanne (EPFL)\\
	Lausanne, Switzerland \\
	\texttt{stefano.zamuner@epfl.ch} \\
	\And
	Paolo De Los Rios \\
	Laboratory of Statistical Biophysics, Institute of Physics, \\
	School of Basic Sciences and Institute of Bioengineering, \\
	School of Life Sciences, \\
	École Polytechnique Féderale de Lausanne (EPFL)\\
	Lausanne, Switzerland \\
	\texttt{paolo.delosrios.epfl.ch} \\
}

\begin{document}
\maketitle

\newcommand{\W}[1]{W^{[#1]}}
\newcommand{\B}[1]{b^{[#1]}}
\newcommand{\x}{\mathbf{x}}
\newcommand{\y}{\mathbf{y}}
\newcommand{\z}{\mathbf{z}}
\newcommand{\E}{\mathbf{l}}
\newcommand{\convergence}{\stackrel{\mathclap{\normalfont\mbox{conv}}}{\;\;=\;\;}}
\newcommand{\comment}[1]{}

\begin{abstract}
Because of the pervasive usage of Neural Networks in human sensitive applications, their interpretability is becoming an increasingly important topic in machine learning. In this work we introduce a simple way to interpret the output function of a neural network classifier that take as input categorical variables. 
By exploiting a mapping between a neural network classifier and a physical energy model, we show that in these cases each layer of the network, and the logits layer in particular, can be expanded as a sum of terms that account for the contribution to the classification of each input \emph{pattern}.
For instance, at the first order, the expansion considers just the linear relation between input features and output while at the second order pairwise dependencies between input features are also accounted for.
The analysis of the contributions of each pattern, after an appropriate gauge transformation, is presented in two cases where the effectiveness of the method can be appreciated.    
\end{abstract}

\section{Introduction}
The increasing and ubiquitous use of machine learning (ML) algorithms in many technological \cite{BADUE2021113816}, financial \cite{de2018machine,ghoddusi2019machine} and medical applications \cite{wang2019deep} calls for an improved understanding of their inner working that is, calls for more \textit{interpretable} algorithms. 
Indeed difficulties in understanding how neural networks operate constitute a major problem in sensitive applications such as self-driving vehicles or medical diagnosis, where errors from the machine could result in otherwise avoidable accidents and human losses. 
Actually, the impossibility to fully grasp the decision process undertaken by the network not only prevents humans from being able to supervise such decision and eventually correct it, but also hinders our ability to use these algorithms to better understand the problem under scrutiny, and to inspire new improved methods and approaches for solving it. 
Thus, the development and deployment of interpretable neural networks could represent an important step to improve the user trust and consequently to foster the adoption of Artificial Intelligence systems in common, everyday tasks \cite{8591322,10.1145/3278721.3278776}. 

Although the concept of \emph{interpretability} is somehow vague \cite{doshivelez2017rigorous}, many methods have been developed with the aim of highlighting how the features of the input directly affect the output of the network \cite{10.1145/3359786}.
In broad strokes, two classes of methods are currently used to analyse trained networks:
\begin{itemize}
\item \emph{Back-propagation-based methods} calculate the gradient of the output with respect to a specific input feature using back-propagation in order to derive the contribution of such feature \cite{Simonyan2014DeepIC,DB15a,bach2015pixel}.
\item \emph{Perturbation explanation methods} are based on the assumption that machine learning predictions in proximity of a given input can be approximated by a linear model of the input features \cite{ribeiro2016why,NIPS2017_0060ef47}. The contribution of a feature can be determined by measuring how the prediction score changes when the feature is altered.
\end{itemize}

Such methods are focused on understanding how the network processes a specific input, and therefore provide a \emph{local interpretation} of the network.
On the contrary in this paper we attempt to provide a \emph{global} description of the output of the neural network, which highlight how the input is processed to produce the output.
In order to achieve our goal we restrict our focus to neural network classifiers that take as input only categorical features and that process them through $\mathcal{C}^\infty$ activation functions.

The global function describing the output of such classifiers can be written as a sum of terms which account for the contribution of all possible combination of input features to the output.
The coefficients that multiply each term can be computed both as derivatives of the output with respect to its input features (similar to gradient based methods), or by feeding the network with artificial input data and measuring how the output is affected (similar to perturbation based methods).
We provide a simple implementation (\url{https://gitlab.com/LBS-EPFL/code/lbsNN/-/tree/v3.0}) and exploit it to investigate how neural network classifiers process their inputs in two different applications.

\section{Results}
We represent vectors such as inputs ($\x$), outputs ($\y$) or hidden layers ($\z$) with bold latin letters.
Each entry $x_i^\alpha$ of the input vectors corresponds to a particular category $\alpha$ of the input feature $i$.
The same is true for the output vector $\y$ and the logit vector $\E$, but as we are assuming for simplicity that only one feature is predicted by the network, we will drop the latin subscript and just use $y^\gamma$ and $l^\gamma$ respectively. 
The approach can be easily generalized to multi-feature predictions.

\subsection{Explicit representation of input-output transfer function}
In modern machine learning a classification task requires inferring a conditional probability that some data $\x$ belongs to a certain class $y^\alpha$. This problem is solved by recursively applying a combination of linear and non-linear parametric transformations, often referred to as activation functions, of $\x$ in a multi-step process. This results in a parametric function $\E_\theta(\x):\mathcal{R}^N\rightarrow\mathcal{R}^K$ that maps the real-valued data $\x \in \mathcal{R}^N$ into real numbers called logits.
The final categorical probability distribution can be obtained by applying the softmax activation function to the logits
\begin{equation}\label{eq:classifier}
p(y^\gamma | \x) = \frac{\exp\left(l^\alpha(\x)\right)}{\sum_{\eta} \exp\left(l^\eta(\x)\right) } = \frac{\exp\left(l^\gamma(\x)\right)}{Z^\gamma} = softmax\left( \E(\x) \right).
\end{equation}

The output of each computational step is in general a quite complicated function of the input data.
In case the nonlinear activation used at each layer are continuous and infinitely derivable, a Taylor expansion of any intermediate output can in principle be computed, but would contain an infinite number of terms.

This is not the case if the input data are binary variables $x_i^\alpha\in \{0,1\}$, or can be cast in such a form, for example using the traditional one-hot encoding.
In this case, since each input bit is idempotent (\textit{i.e.} $\left(x_i^\alpha \right)^n = x_i^\alpha$ if $n>0$), any output function can be represented as a polynomial expansion in which each input data appears at most with power one:
\begin{equation}\label{eq:expansion_general}
z_\ell(\x) = W_{\ell}+\sum_i\sum_\alpha W^{\alpha}_{i,\ell}\;x_i^\alpha + \sum_{i,j\neq i}\sum_{\alpha,\beta} W^{\alpha,\beta}_{ij,\ell}\;x_i^\alpha x_j^\beta+\cdots.
\end{equation}
Equation \ref{eq:expansion_general} provides a global and exact function that can express the output of any layer $\z$, for any input $\x$.
Each order of the polynomial describes the contribution of combination of an increasing number of input features to $\z$.
For example the term $W^{\alpha}_{i,\ell}$ represents a \emph{coupling} between $z_\ell$ and $x_i^\alpha$: the higher its absolute value, the stronger will be the influence of $x_i^\alpha$ to the value $z_\ell$. Similarly,
higher order terms capture the \textit{interaction strength} between $z_\ell$ and a combination of multiple input variables.
The degree of the polynomial is at most equal to the number of input features.

\subsection{Gauge fixing}
Up to now we used the fact that each entry in the input vector is either a zero or a one. 
Categorical features have another property that will be relevant in the following: each feature can only be of one category, or $\sum_{\alpha} x_i^\alpha=1$.
The most prominent consequence of this fact is that an infinite number of choices of coefficients in (\ref{eq:expansion_general}) can result in the very same value of $z_\ell(\x)$.
In physics, transformations that change the terms of a function but do not result in a change of the function itself are called \emph{gauge transformations} \cite{RevModPhys.73.663}.
For instance, it is easy to see that the function $z_\ell(\x)$ in Equation~\ref{eq:expansion_general} can also be written as
\begin{equation}
z_\ell(\x) = \left(W_{\ell}+G\right)+\sum_i\sum_\alpha \left(W^{\alpha}_{i,\ell}-\frac{G}{N}\right)\;x_i^\alpha + \sum_{i,j\neq i}\sum_{\alpha,\beta} W^{\alpha,\beta}_{ij,\ell}\;x_i^\alpha x_j^\beta+\cdots,
\end{equation}
for every choice of $G$.

As the parameters of Equation~(\ref{eq:expansion_general}) can vary depending on the choice of the gauge, and with them our interpretation of the relevance of each input features, it is necessary to fix the gauge before interpreting the expansion.

In this paper we chose to use the Ising gauge, \textit{i.e.} to transform every coefficient in the expansion in such a way that the average of the coefficients over all the possible categories of the associated features is zero.
Such transformation for the coefficients $W_{ij,\ell}^{\alpha,\beta}$ reads:
\begin{align}\label{eq:gauge}
&W_{ij,\ell}^{\alpha,\beta}\rightarrow W_{ij,\ell}^{\alpha,\beta} - \frac{1}{n_j}\sum_\beta W_{ij,\ell}^{\alpha,\beta} - \frac{1}{n_i}\sum_\alpha W_{ij,\ell}^{\alpha,\beta} + \frac{1}{n_i n_j}\sum_{\alpha,\beta} W_{ij,\ell}^{\alpha,\beta}\\
&W_{i,\ell}^{\alpha}\rightarrow W_{i,\ell}^{\alpha} + \sum_j\frac{1}{n_j}\sum_\beta W_{ij,\ell}^{\alpha,\beta}\\
&W_\ell\rightarrow W_\ell+\sum_{i,j\neq i}\frac{1}{n_i}\sum_{\alpha,\beta} W_{ij,\ell}^{\alpha,\beta}-\sum_{i,j\neq i} \frac{1}{n_i n_j}\sum_{\alpha,\beta} W_{ij,\ell}^{\alpha,\beta}
\end{align}
Similar equations can trivially be derived for all orders of the polynomial in Equation~\ref{eq:expansion_general}.

As exemplified by Equations~\ref{eq:gauge}, imposing a gauge to the terms of a given order generates a cascade of terms that modify all lower orders of the expansion. To correctly fix the gauge it is therefore important, in theory, to gauge-transform the coefficients of the expansion starting from the higher order ones. In practice we found that it is usually sufficient to fix the gauge of the lower orders of the expansion, as the additional terms that comes from higher orders have little impact on the coefficients of the lower ones. In this paper, for example, all results have been obtained by fixing the gauge of the second and first order of the expansion, but not that of higher orders.

The net effect of applying the Ising gauge is that the contributions of higher order terms is reduced in favor of equivalent lower order contributions: the role of lower order patterns is therefore emphasized.

\subsection{Physical interpretation of the logits layer}
Classifiers such as (\ref{eq:classifier}) can be interpreted as energy models \cite{grathwohl2020classifier} where the logits play the role of the joint energy $E(\x,y^\ell)$ between input data $\x$ and output category, i.e. $l^\ell(\x)=-E(\x,y^\ell)$.
When we consider the Taylor expansion of the logits layer, we can interpret the terms of the expansion as the contributions of different features, or group of features, to the joint energy between input and output.
In this interpretation $x_i$ and $y$ can be seen as physical objects that change together because of the soft constraints imposed by their joint energy.
For instance, the coefficients $W_{i,\ell}^{\alpha}$ can be interpreted as couplings between the output category $y^\ell$ and the input feature $x^\alpha_i$. Similarly, $W_{ij,\ell}^{\alpha,\beta}$ describe the contributions of the \emph{three-body interactions} between inputs $x^\alpha_i$,$x^\beta_j$  and output $y^\ell$.

The functional form of the energy of such a classifier is reminiscent of that of a Potts model \cite{RevModPhys.54.235}, where not only pairwise couplings contribute to the energy, but also higher order interactions can be important.
In the case of a neural network classifier, all possible interactions (up to $N$-body interactions) contribute to the joint energy $E(\x,y^\ell)$, although, as mentioned above, the Ising gauge emphasizes, if possible, the role of the lower-order interactions.  

\subsection{Maximum entropy interpretation of the logits layer}

When the network is trained to minimize the categorical cross-entropy $\mathcal{L}(y,\hat{y})$ between the expected output $\tilde{\y}$ and the predicted one $\y$, we can observe that
\begin{equation}\label{eq:loss_derivative}
\frac{d\langle\mathcal{L}(y,\tilde{y})\rangle}{d\;W_{i,\ell}^\alpha}=\langle \tilde{y}^{\ell}\;x_i^\alpha \rangle - \langle y^\ell x_i^\alpha \rangle \convergence 0
\end{equation}
where $\langle \cdot \rangle$ is the average over all training samples, and the last equality is true at convergence.
The derivative in Equation \ref{eq:loss_derivative} can be computed starting from the derivatives of the loss function with respect to the parameters of the network usually computed by back-propagation and by employing the chain rule.
When the training reaches convergence, the derivative of the loss function $\mathcal{L}$ with respect to any parameter must be zero, implying that the trained network reproduces the input-output correlations present in the training set.

The functional form of the joint energy, together with the constraints described in Equation \ref{eq:loss_derivative}, suggest a maximum-entropy \cite{PhysRev.106.620,PhysRev.108.171} interpretation of the output of the classifier.
It is indeed easy to show that the maximum of the functional
\begin{equation}
    F(\x,\y) = -\sum_\ell p(\x,y^\ell)\;\log\left( p(\x,y^\ell) \right) - \sum_\ell\; W_\ell \langle y^\ell \rangle - \sum_\ell\sum_{i,\alpha}\; W_{i,\ell}^\alpha \langle y^\ell\; x_i^\alpha \rangle - \cdots, 
\end{equation}
which is the probability distribution that have the highest entropy while also satisfying all constraints from Equation \ref{eq:loss_derivative}, precisely takes the functional form from Equation \ref{eq:classifier} with

\begin{equation}\label{eq:expansion_logits}
l^\ell(\x) = W_{\ell}+\sum_i\sum_\alpha W^{\alpha}_{i,\ell}\;x_i^\alpha + \sum_{i,j\neq i}\sum_{\alpha,\beta} W^{\alpha,\beta}_{ij,\ell}\;x_i^\alpha x_j^\beta+\cdots.
\end{equation}

We therefore can interpret each coefficient $W$ of the Taylor expansion of the logits layer as a Lagrange multiplier that constrains the corresponding correlation between input and output in the training set.

\subsection{Coefficients of the expansion}
The functional form  (\ref{eq:expansion_general}) suggests two possible ways for computing the coefficients of the expansion:
\begin{itemize}
\item \emph{As a derivative} of the layer of interest with respect to all the corresponding input feature(s). For instance 
\begin{equation}
W_{ij,\ell}^{\alpha,\beta} = \left.\frac{d z^\ell(\x)}{ dx_i^\alpha\;dx_j^\beta }\right|_{\x=\mathbf{0}}.
\end{equation}
This is reminiscent of all derivative method for interpreting neural networks, with the important difference that such derivatives are computed at the invalid input $\x=\mathbf{0}$.
\item \emph{Computing the output of the layer of interest} on different synthetic input vectors. For instance $W_\ell=z^\ell(\mathbf{0})$, whereas $W_{i,\ell}^\alpha = z^\ell(\mathbf{\delta}(x_i^\alpha))-z^\ell(\mathbf{0})$, where $\delta(x_i^\alpha)$ is the vector with a unique non-zero entry in position $x_i^\alpha$. In practice we can reconstruct the coefficients of the expansion by repeatedly perturb the artificial input $\x=\mathbf{0}$ and testing how the output is affected by such perturbations.
\end{itemize}

It is important to notice that even if the way the coefficients are computed is similar to what is done with other techniques, the result is quite different.
Indeed, perturbation and derivative methods generally probe how the output of the layer of interest responds to perturbations of a given input vector and try to describe how that particular input vector is processed by the network.
In our method we try to assess how the network processes all possible inputs by perturbing (or computing derivatives) in proximity of a non valid input vector $\mathbf{0}$.
In order to obtain the complete expansion of the network we would have to effectively compute its output for every possible input, which is infeasible in most cases.\newline
In the following examples we will show that the low order terms are the dominant ones, with the higher orders only marginally contributing to the final output of the network. This will allow us to neglect the higher orders of the expansion without losing relevant information on how the input are processed by the network.
In some applications such as Direct Coupling Analysis (our second example) the fact that low order terms dominate the expansion is to be expected on physical grounds.

\subsection{Examples}\label{sec:examples}

In the first example we trained a shallow network to recognize digits from a two and three bit versions of the MNIST dataset \cite{lecun-mnisthandwrittendigit-2010}.
In the second example we trained several shallow networks, each one aimed to predict one amino acid in a protein sequence using as input all the remaining ones. 
This second example is devised in a way to mimic existing techniques of Direct Coupling Analysis (DCA) \cite{morcos2011direct} that are used to predict interactions between amino acid pairs, while extending them to consider higher order interactions.

In order to display the results of our technique we made the assumption that the lower order terms in the expansion of the logits layer are the most important.
In particular we considered the expansion of the network just up to the second order in the input features, and consequently applied the gauge transformation only to first and second order terms.
We could expect this to be a limitation in the MNIST classifier where patterns composed of multiple pixels are clearly visible, but not very relevant for the DCA example where the physical interactions we are trying to capture typically involve a limited number of amino-acids.

Moreover, although visualizing the coefficients of the first two orders of the expansion is still possible, we found that it is often clearer to display a score that describes the general effect of $x_i$ for the output without focusing on the effect of each specific category, $x_i^\alpha$.
We therefore define the scores $S_{i,\ell}=\sum_\alpha \left(W_{i,\ell}^\alpha\right)^2$ and $S_{ij,\ell}=\sum_{\alpha,\beta} \left( W_{ij,\ell}^{\alpha,\beta}\right)^2$, i.e. the Frobenius' norm of all the coefficients that couple the output $y^\ell$ to the input features $x_i$ and $x_j$.

It is worth reminding that the application of the Ising gauge results in the lowest possible scores for the higher order patterns, progressively shifting the weights to the lower order ones, thus using simpler terms in the network interpretation and visualization.

\subsubsection{MNIST dataset}
When training on the MNIST dataset, the coefficients $W_{i,\ell}^\alpha$ estimate the contribution of the value $\alpha$ in pixel $i$ of the input image on the probability that that image is classified as $\ell$. 
Similarly, the score $S_{i,\ell}$ will summarize the effect of pixel $i$ on the discriminator output independently from the value the pixel assumes: a zero score is obtained when the pixel does not affect the probability of $y^\ell$ while a high score means that the pixel is important for determining that probability.

\begin{figure}[h!]
\begin{center}
\includegraphics[width=0.8\textwidth]{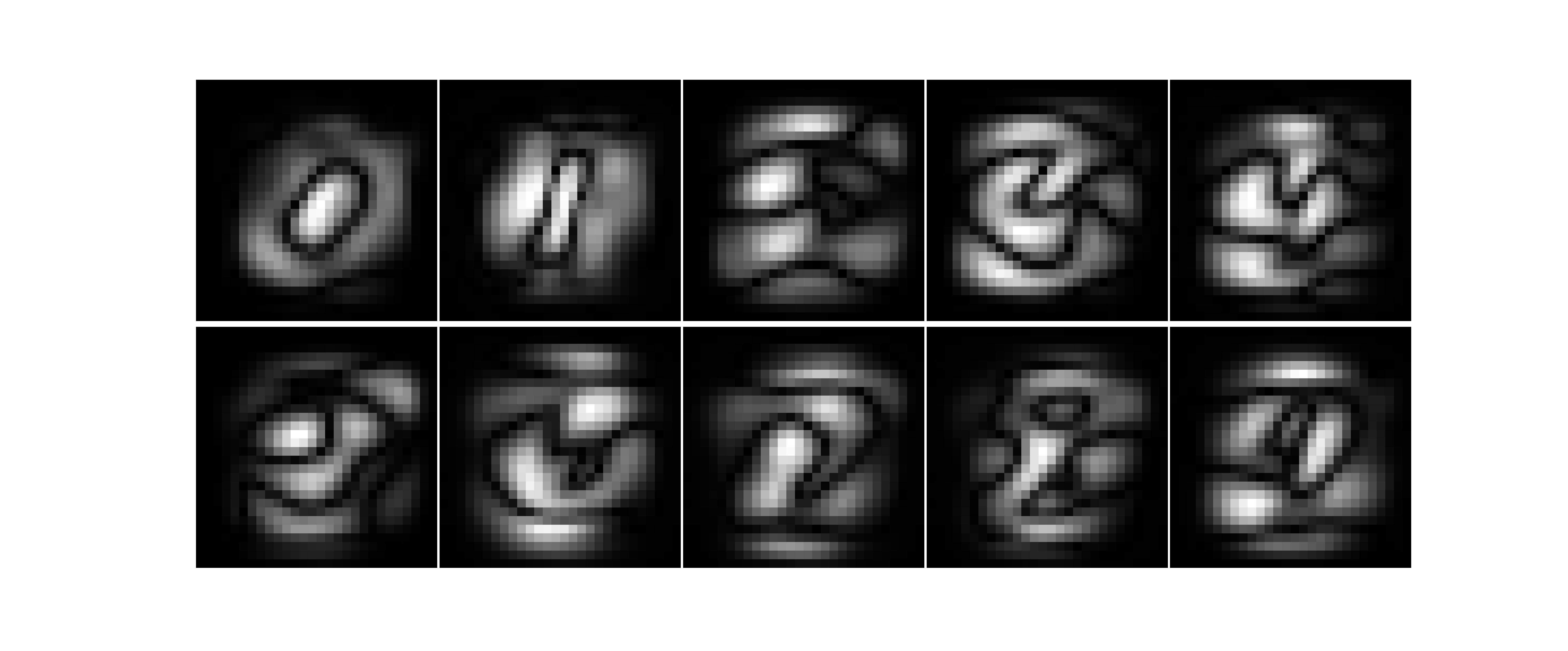}
\caption{Scores from first order expansion of the Energy learned from our 2-bits MNIST dataset. Black corresponds to zero. Features that resembles the corresponding digits are clearly visible}
\label{fig:2bits_score1}
\end{center}
\end{figure}

Figure \ref{fig:2bits_score1} shows the first order scores $S_{i,\ell}$ obtained on our 2-bits MNIST dataset for every output category.
It is clear that, at the first order, our shallow network recognizes patterns with some resemblance to the actual digits.

\begin{figure}[h!]
\begin{center}
\includegraphics[width=0.8\textwidth]{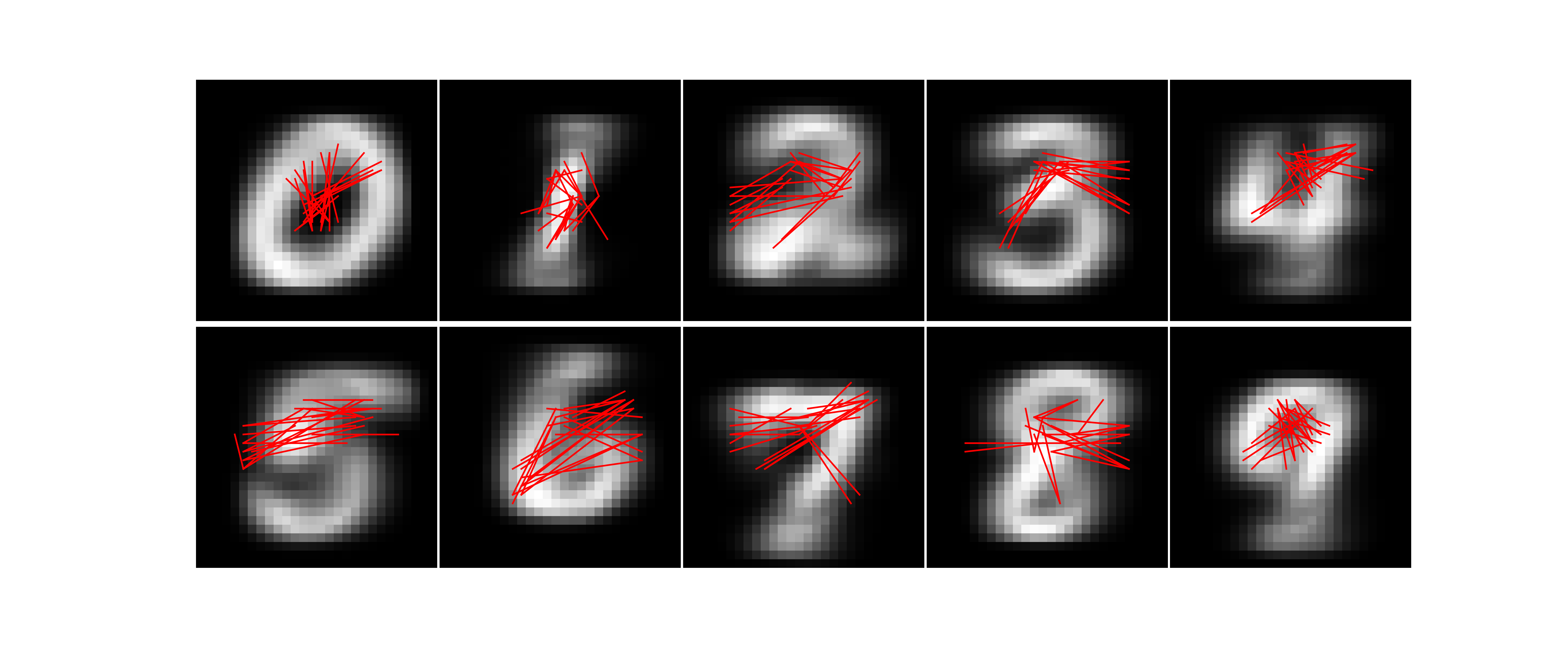}
\caption{Visualizations of the second order expansion of our discriminator on our 2-bits MNIST dataset (red lines) on top of average digits in out training dataset (grayscale). Pixels connected by the same line constitute important pairs in the expansion.}
\label{fig:2bits_score2_lines}
\end{center}
\end{figure}

At the second order of the expansion, the scores measure the joint contributions of two pixels for the final classification. This information is in general more difficult to represent and interpret. In Figure \ref{fig:2bits_score2_lines} the $20$ non-adjacent pixel pairs that exhibit the highest second order score are connected by a red line.

Whereas the first order scores describe the importance of each pixel independently from all the others, the second order score highlights the importance of pixel pairwise correlations (both pixels with similar color) or anticorrelations (pixels with dissimilar colors). 
For instance, if we focus our attention on the \textit{zero} digit in Figure \ref{fig:2bits_score2_lines}, we see that the first order terms recognize the relevance of pixels belonging to the symbol of the number $0$ or to its inner area, which are the fundamental characteristics of such digit. Yet, they do not bring any information about their mutual relation. The most important scores in second order interactions, instead, highlight that the symbol $0$ (on average white) and its internal void (on average black) are anticorrelated. The network is thus recognizing the importance of specific written and unwritten regions, and the way they are related: in the case of the digit $0$, this means a central void surrounded by white pixels. 
Similar results are obtained using a 4-bits version of the dataset.

To estimate the relevance of each term of the expansion, we tried to classify each image in the training set using the whole network, all terms up to the second order, or all terms only up to the first order. 
Performance degradation after the removal of one order is indicative of the importance of that order of the expansion.

\begin{table}
\begin{center}
\begin{tabular}{rcc}

Network description & accuracy on training set & accuracy on test set \\ \hline
 whole network & $0.948$ & 0.914 \\
 second order expansion & 0.901 & 0.908 \\
 first order expansion & 0.705 & 0.725
\end{tabular}
\end{center}
\caption{Accuracy of the network when removing higher order terms in the expansion compared to the accuracy of the whole network.}
\end{table}

In this case the performance of the truncated network output, with up to second order terms, is similar to the one of the whole trained network.
A considerable performance degradation is instead obtained when the second order terms are not used for the predictions.
This suggest that in this case analyzing the first two orders of the expansion is sufficient to describe reasonably well the behavior of the network, with higher order terms in charge of accounting the remaining details. This justify \emph{a posteriori} our choice of neglecting higher order terms. 

\subsubsection{Direct Coupling Analysis on RAS protein family}
Unveiling the relationship between protein sequence and protein structure is a long standing problem in biology.
Direct Coupling Analysis (DCA) is an umbrella term for many bioinformatics methods that have gained a lot of attention in recent years for their ability to infer information about the shared structure of homologous proteins by using their amino-acid sequences as input data \cite{coucke2016direct,morcos2011direct,muscat2020filterdca,barrat2020sparse,jacquin2016benchmarking,muntoni2020aligning}.
DCA takes as input a number of aligned amino-acid sequences of length $N$ and returns as output an $N \times N$ matrix of pairwise scores between amino-acids: high scores are indicative of an important mutual evolutionary influence between amino-acids, while small scores suggest a relative evolutionary independence.
As such influence must be mediated by physical interactions, high scored pairs are usually in close proximity in space.
In order to assess the quality of DCA prediction we can therefore compare the highest predicted scores to the NxN matrix containing all pairwise distances (distogram) of a known structure of the same protein.
When the number of sequences is sufficiently large, the predictions from DCA are typically highly accurate.

The success of these approaches is testified by their use as prominent input features in many deep learning algorithms that have reached outstanding performances in protein structure prediction such as AlphaFold \cite{senior2020improved} and trRosetta \cite{yang2020improved}.

Among all DCA methods, Pseudolikelihood DCA (plmDCA) \cite{PhysRevE.87.012707} is often employed because it offers reasonable computational speed  without compromising the accuracy of its predictions.
In plmDCA the inference proceeds by iteratively training a linear classifier to infer the amino-acid type occupying each position of the sequence, given the amino-acid types in all the other positions.
The final scores are finally computed from the coefficients of the linear classifier using the Frobenius norm as described in Section \ref{sec:examples}.

In this example we substitute the linear classifiers with deeper, but still shallow, neural networks. 
We engineered the networks in such a way that all the terms of order three or larger in the expansion would be zero by using a custom activation function that simply squares the output of a single hidden layer. In this case therefore, the second order expansion of the network is not an approximation and the gauge can be fixed exactly.
We trained such networks in order to infer information about the structure of a small GTPase protein involved in cell-growth known as RAS \cite{harvey1964unidentified}. RAS is an important and well studied protein as mutations in the Ras genes can ultimately lead to cancer in humans \cite{downward2003targeting,malumbres2003ras}.

In this context, the coefficients of the first order of the expansion of each network $W^{\alpha,[k]}_{i,\ell}$ (the suffix in square brackets denote the missing residue we are trying to predict), can be interpreted as the interactions between the amino-acid in position $i$ and the predicted one $k$.
Similarly, the coefficients of the second order terms can be seen as three-body interactions between the missing amino-acid and two others of the same sequence.

\begin{figure}[h!]
\begin{center}
\includegraphics[width=0.95\textwidth]{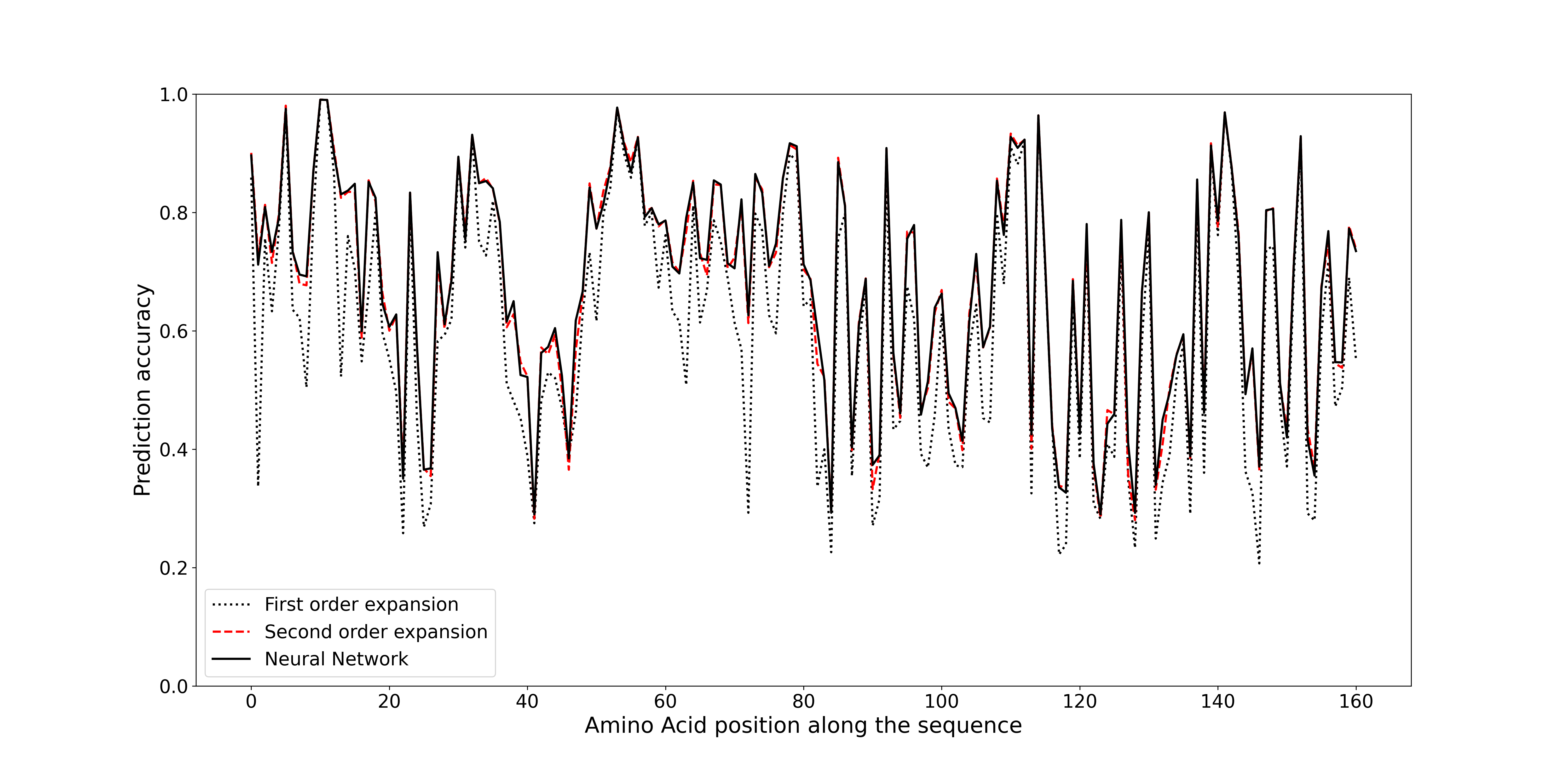}
\caption{Prediction accuracy for every amino acid in RAS sequence. Using the whole network (solid black lines) provide only marginal improvement over its first order expansion (dotted line) in most cases. Differences between the performance of the whole network and its second order expansion (red line) are to be imputed to numerical errors during the estimation of the coefficients. }
\label{fig:lbsnn_ras_performance}
\end{center}
\end{figure}

As shown in Figure \ref{fig:lbsnn_ras_performance}, in most cases the performance of a classifier that only contains the first order expansion of the network closely reproduces the performance of the whole network. There are nonetheless  several cases for which the addition of second order terms provide a significant improvement.

The presence of such multi-body interactions could be imputed to physical interactions, as it is typically the case for the two-body ones, but also to effective interactions due to the presence of an external interacting object, such as a ligand. We therefore expect the amino acids involved in such multi-body interactions to be spatially close and possibly close to binding sites.

\begin{figure}[h!]\label{fig:lbsnn_ras_map}
\begin{center}
\includegraphics[width=0.45\textwidth]{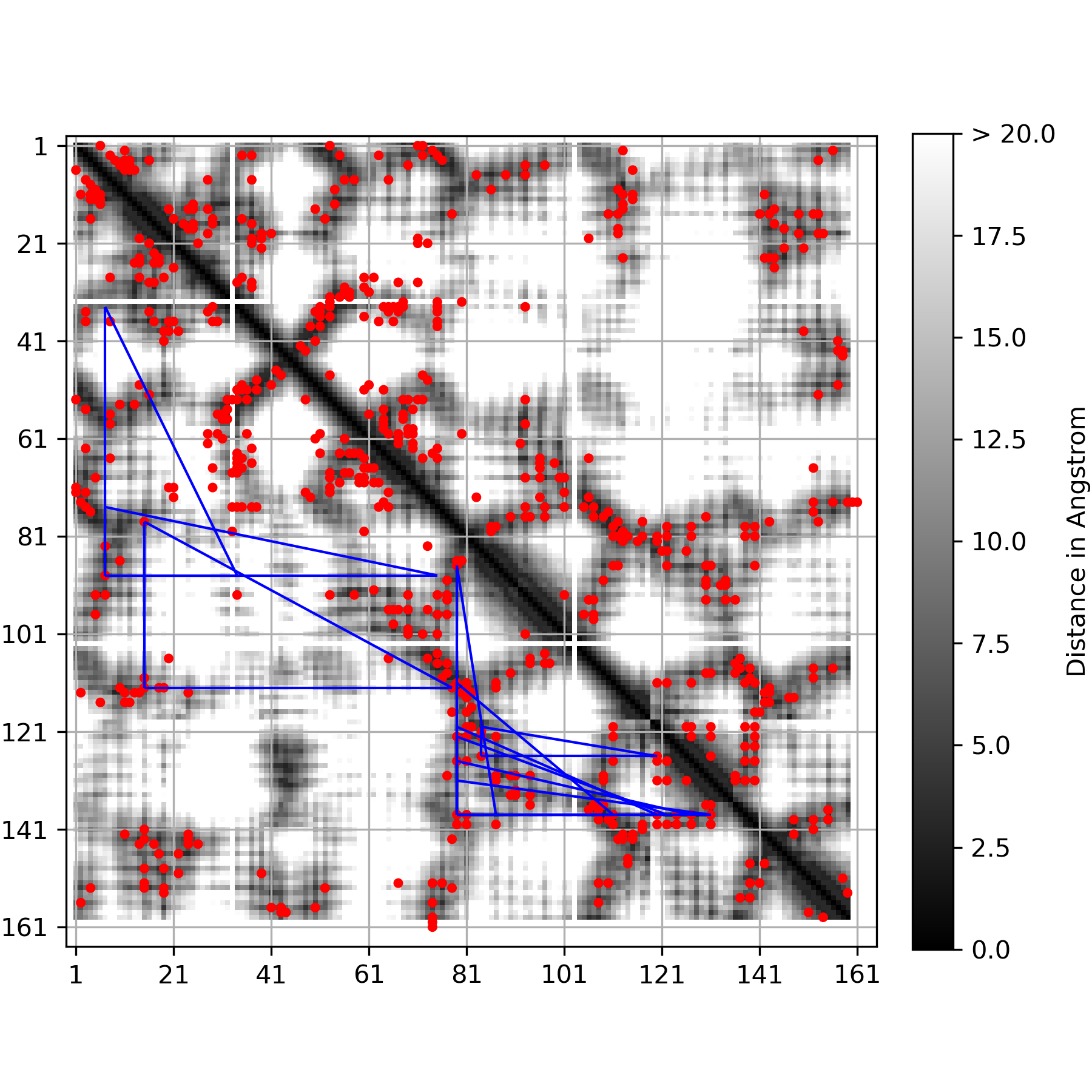}
\quad
\includegraphics[width=0.45\textwidth]{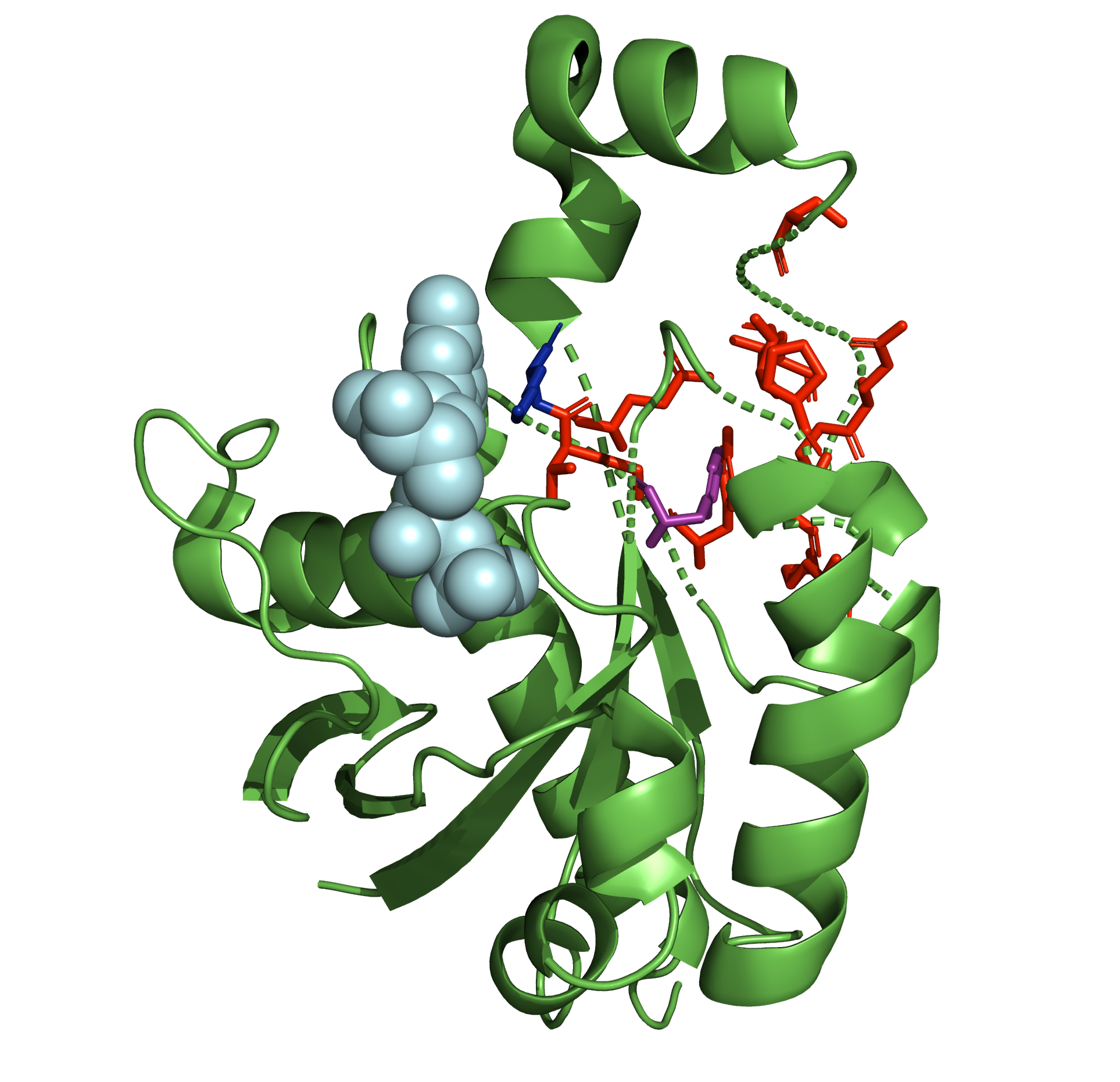}
\caption{First and second order of the expansion of our neural networks trained on RAS protein family. Left panel: comparison between top 322 highest scored pairs (red dots) using our method and the observed distances in experimental structure 5vcu (shown in grayscale). Predictions between pairs $(i,j)$ with $|j-i|<5$ are not shown for visualization purposes. The 10 most important three-body interactions are here visualized with blue triangles where each vertex correspond to a pair in the triplet. Right panel: cartoon representation of RAS protein (green) bounded to GDP (light blue spheres) from pdb 5vcu. Residues involved in the 10 most important triplets are shown as sticks: residue K116 (dark blue) interacts with GDP, residue F82 from the \emph{effector lobe} is shown in purple, all other residues (P87,L90,G114,T115,S135,V137,T139,Q141,G142,K146,Y154,E156) in red. See also \ref{fig:lbsnn_ras_struct_2} \label{fig:lbsnn_ras_struct}  }
\end{center}
\end{figure}

\begin{figure}[h!]
\begin{center}
\includegraphics[width=0.80\textwidth]{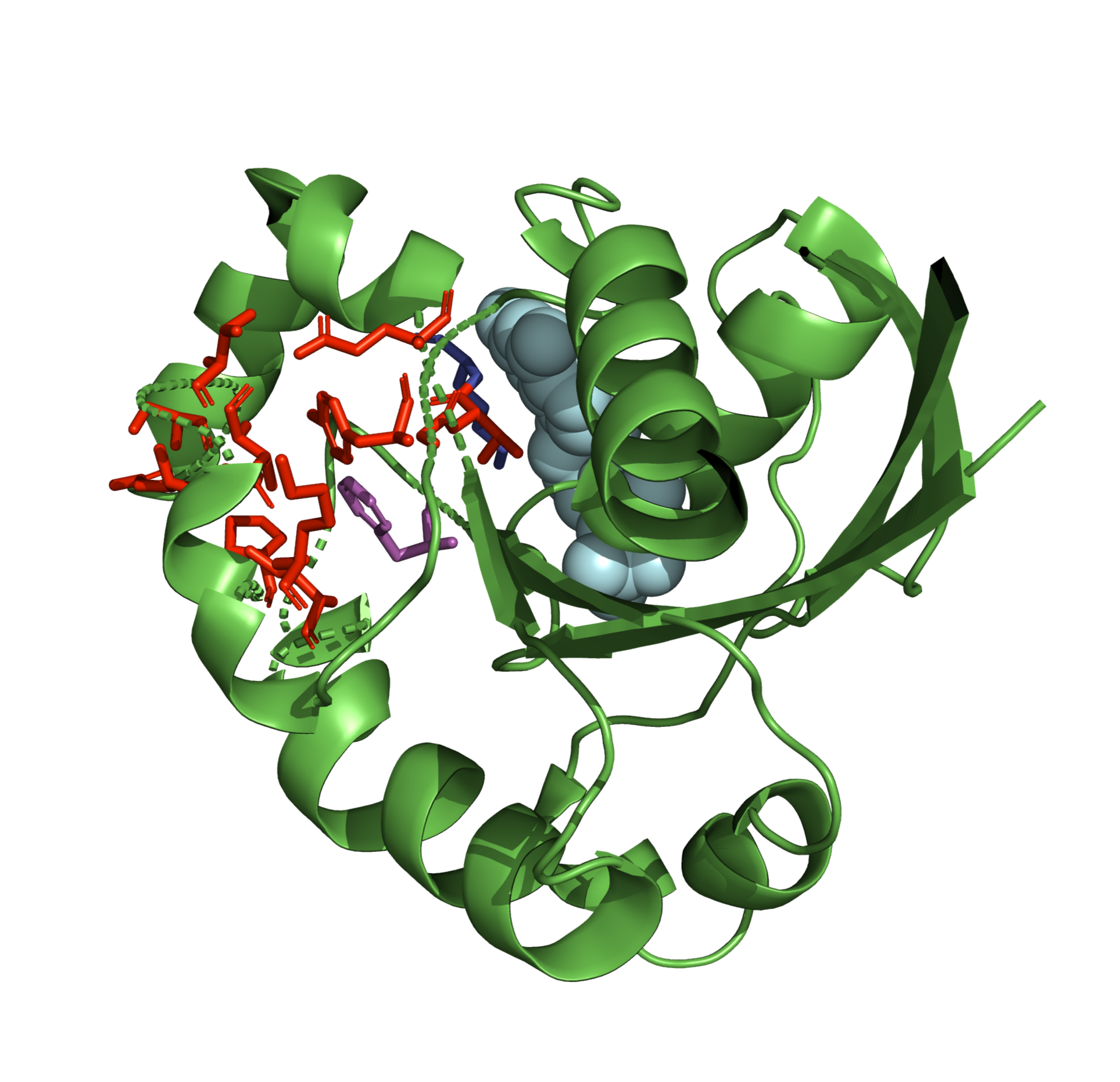}
\caption{Alternative view of the structure of RAS (pdb code: 5vcu) highlighting all amino acids involved in the top scored triplets. Light blue spheres: GDP. Purple sticks: residue F82. Blue sticks: residue K116, interacting with GDP. All other residues involved in the 10 most important triplets are displayed as red sticks. \label{fig:lbsnn_ras_struct_2}}
\end{center}
\end{figure}

The left panel in Figure \ref{fig:lbsnn_ras_struct} shows the 322 most relevant pairs in the expansion of the logits layer (red dots) superposed to the distance map of the experimental structure of RAS (pdb code: 5vcu).
Most predicted pairs corresponds to amino-acid pairs that are close in the three-dimensional structure, as hypothesized. 
Similar predictions can be obtained using standard DCA techniques (see Figure \ref{fig:lbsdca_ras_map}) but our method also allows to determine which are the most relevant three-body interactions, shown here as blue triangles connecting all the pairs inside each triplet.
The spatial distribution of the residues involved in the 10 most prominent triplets is displayed in the right panel of Figure \ref{fig:lbsnn_ras_struct}, which shows a cartoon representation of the Ras structure from pdb 5vcu. 
Among these, residue K116 (in dark blue sticks), interacts with the nucleotide ligand of RAS (light blue, represented as spheres). Interestingly, whereas most residues from important triplets belong to the \emph{allosteric lobe} of the protein \cite{lu2016ras} (which contains hot spots of protein–ligand interactions that bind membrane components), only one (residue F82, shown as purple sticks) is found in the \emph{effector lobe}.
All other residues involved in three body interactions are displayed as red sticks. See also Figure~\ref{fig:lbsnn_ras_struct_2} for an alternative view of the structure.

\section{Conclusions}
We introduced a novel technique for interpreting the output of Neural Network classifiers that take as input categorical variables.
By taking advantage of the mathematical properties of such inputs, we showed that the output of each layer of the network can be written as a power series with a finite number of simple terms.
In particular, this allows to obtain a global function that approximate the output of the network with any required degree of accuracy.
We took advantage of a mapping between energy models and classifiers to interpret each terms of the expansion of the logits layer as the relevant \emph{interactions} between combinations of input features and the output.
Similarly, we recognized that the functional form of the expanded logits exactly matches the one obtained by a maximum entropy model where the correlations between combinations of input features and the output are constrained.
In this perspective, the network seems to produce the output that best matches the most relevant correlations with the input features it has learned during the training phase.

In order to assess the method, we applied it to two different contexts: recognition of handwritten digits in the MNIST dataset and prediction of amino-acids in RAS protein sequences.
Unexpectedly, we proved that, when predicting digits from the MNIST dataset, the two lower orders of the expansion can account for most of the information used by the network. 
In many cases, when predicting amino-acids from a protein sequence, the first order term of the expansion was sufficient for producing correct predictions. However, in several cases the second order of the expansion led to a significant improvement of the predictions.\newline
Residues belonging to the most relevant predicted triplets are clustered in the structure of the protein, and all belong to regions involved in the binding with nucleotides or other ligands that bind membrane elements. These results suggest that three-body interactions have a role in the evolution of protein sequences and that their study can provide hints about the functioning of the protein. We plan to explore the application of this technique to the study of protein sequences in a future paper.

Although this method is exact only in this restricted application cases, the same ideas can be used to provide global approximated function to neural networks classifiers with real valued inputs.
In such cases, the expansion would contain an infinite number of terms and therefore only approximated representations of the output of the network could be obtained. Moreover, in contrast to what happens with categorical inputs, each input would appear in the expansion with all integer powers, making it in principle impossible to describe the contribution of each input feature to the output. Nonetheless, a reduced representation in which only the first orders of the expansion are taken into account could be obtained. \newline
The presence of high order terms in the description of the output could be, supposedly, problematic as small perturbations of the corresponding inputs could results in disruptive effects to the output of the network, making it exploitable through adversarial attacks.
We could therefore speculate that, by downplaying the role of high powers of input features, the approximate network could be more robust to adversarial attacks. Moreover, as shown in the proposed examples, such reduced description is not necessarily less accurate than the whole network.

\section{Methods}
All neural networks have been implemented in Tensorflow2 \cite{tensorflow2015-whitepaper}.
Code are data are made available at \url{https://gitlab.com/LBS-EPFL/code/lbsNN/-/tree/v3.0} and \url{https://gitlab.com/LBS-EPFL/data/interpretablenndata}.

\paragraph{MNIST}
We used 50000 randomly chosen images from MNIST as training set, and the remaining ones as validation set.
For each pixel position, we computed the mean on the whole training set; we subsequently assigned value 0 to all pixels with value less or equal to the corresponding average, and 1 otherwise.
No data augmentation technique has been employed during training.
We trained a neural network constituted of one hidden layer of 800 units and tanh activation, as the one employed in \cite{simard2003best}. Other architectures have been tested with consistent results.
We trained using Adam optimizer, with learning rate equal to 0.005 and L2 kernel and bias regularizers with coefficient $0.01$.
The final accuracy of the network was 91.4\% on the validation set.

We repeated the experiment with a dataset where each pixel has been mapped to 4 possible values. 
We did this by equally dividing the range of possible values of each pixels  in balanced ranges and then assigning each pixel a number corresponding to  the range it belongs to. Example images showing the 4bits digits are shown in Supplementary Material.
The accuracy of the network increased to 92.3\% in the validation set.
First and second order expansions using the 4 bits dataset are shown in Supplementary Material.

\begin{figure}[h!]\label{fig:lbsdca_ras_map}
\begin{center}
\includegraphics[width=0.80\textwidth]{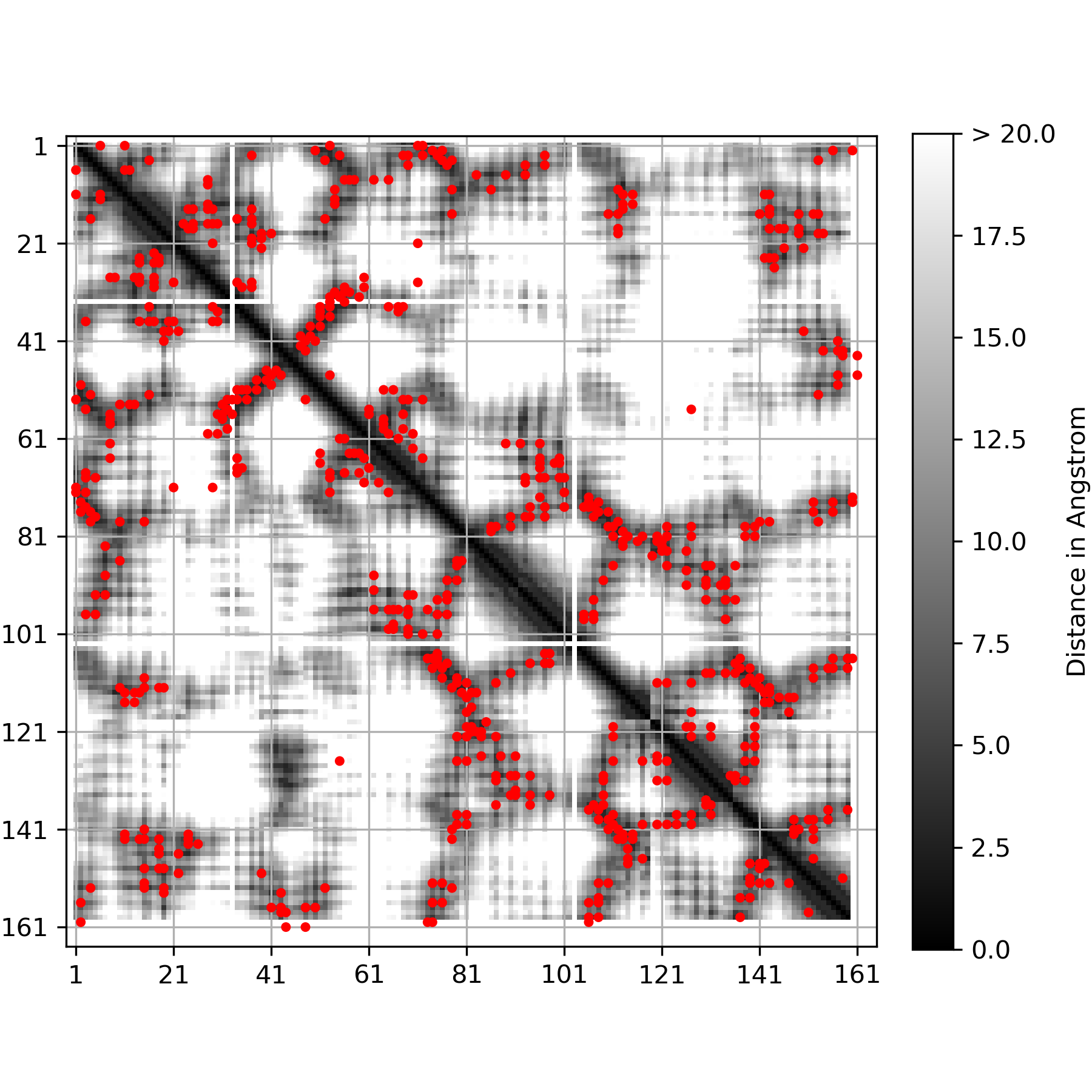}
\caption{Comparison between top 320 highest scored pairs (red dots) as obtained with lbsDCA \cite{malinverni2019coevolutionary} and the observed contacts in a real structure. Gray map: pairwise distances in reference structure (pdb code: 5vcu). Pairwise predictions are comparable with those obtained with our technique, but in this case we cannot infer three body interactions. }
\end{center}
\end{figure}

\paragraph{Ras sequences}
A multiple sequence alignment of Ras (PF00071) has been downloaded from Pfam \cite{bateman2004pfam} on 2020-09-12, together with the corresponding hidden Markov model.
The alignment, initially composed of 78021 sequences, has been reduced to 57947 sequences after removal of sequences with more than $10$\% of gaps.
For each sequence we computed its sample weight as the reciprocal of the number of sequences that share more than $80$\% sequence identity with it.
The same weights have been used as sample weights in both plmDCA and in our shallow network.
Pseudolikelihood DCA has been computed by using lbsDCA (\url{https://gitlab.com/LBS-EPFL/code/lbsDCA/-/tree/v1.0}) \cite{malinverni2019coevolutionary} with default parameters.
Predictions from plmDCA are shown in Figure \ref{fig:lbsdca_ras_map}.

In our shallow network we opted for a single hidden layer with $32$ units and square activation function and L2 kernel regularizers with coefficient $0.01$.
The network has been trained with Adam optimizer and learning rate equal to $0.01$ for $500$ epochs.
Note that, to better compare to plmDCA, the same architecture and hyperparameters have been employed to predict all amino acids of the sequence: we expect that adapting the architecture of the network to each position of the alignment could improve the predictions.
Also, similarly to what is done in plmDCA, we trained the network on the whole sequence alignment.

\bibliographystyle{unsrt}
\bibliography{biblio}

\begin{thebibliography}{10}

\bibitem{BADUE2021113816}
Claudine Badue, Rânik Guidolini, Raphael~Vivacqua Carneiro, Pedro Azevedo,
  Vinicius~B. Cardoso, Avelino Forechi, Luan Jesus, Rodrigo Berriel, Thiago~M.
  Paixão, Filipe Mutz, Lucas {de Paula Veronese}, Thiago Oliveira-Santos, and
  Alberto~F. {De Souza}.
\newblock Self-driving cars: A survey.
\newblock {\em Expert Systems with Applications}, 165:113816, 2021.

\bibitem{de2018machine}
Jan De~Spiegeleer, Dilip~B Madan, Sofie Reyners, and Wim Schoutens.
\newblock Machine learning for quantitative finance: fast derivative pricing,
  hedging and fitting.
\newblock {\em Quantitative Finance}, 18(10):1635--1643, 2018.

\bibitem{ghoddusi2019machine}
Hamed Ghoddusi, Germ{\'a}n~G Creamer, and Nima Rafizadeh.
\newblock Machine learning in energy economics and finance: A review.
\newblock {\em Energy Economics}, 81:709--727, 2019.

\bibitem{wang2019deep}
Fei Wang, Lawrence~Peter Casalino, and Dhruv Khullar.
\newblock Deep learning in medicine—promise, progress, and challenges.
\newblock {\em JAMA internal medicine}, 179(3):293--294, 2019.

\bibitem{8591322}
K.~{Amarasinghe} and M.~{Manic}.
\newblock Improving user trust on deep neural networks based intrusion
  detection systems.
\newblock In {\em IECON 2018 - 44th Annual Conference of the IEEE Industrial
  Electronics Society}, pages 3262--3268, 2018.

\bibitem{10.1145/3278721.3278776}
Rahul Iyer, Yuezhang Li, Huao Li, Michael Lewis, Ramitha Sundar, and Katia
  Sycara.
\newblock Transparency and explanation in deep reinforcement learning neural
  networks.
\newblock In {\em Proceedings of the 2018 AAAI/ACM Conference on AI, Ethics,
  and Society}, AIES '18, page 144–150, New York, NY, USA, 2018. Association
  for Computing Machinery.

\bibitem{doshivelez2017rigorous}
Finale Doshi-Velez and Been Kim.
\newblock Towards a rigorous science of interpretable machine learning, 2017.

\bibitem{10.1145/3359786}
Mengnan Du, Ninghao Liu, and Xia Hu.
\newblock Techniques for interpretable machine learning.
\newblock {\em Commun. ACM}, 63(1):68–77, December 2019.

\bibitem{Simonyan2014DeepIC}
K.~Simonyan, A.~Vedaldi, and Andrew Zisserman.
\newblock Deep inside convolutional networks: Visualising image classification
  models and saliency maps.
\newblock {\em CoRR}, abs/1312.6034, 2014.

\bibitem{DB15a}
J.T. Springenberg, A.~Dosovitskiy, T.~Brox, and M.~Riedmiller.
\newblock Striving for simplicity: The all convolutional net.
\newblock In {\em ICLR (workshop track)}, 2015.

\bibitem{bach2015pixel}
Sebastian Bach, Alexander Binder, Gr{\'e}goire Montavon, Frederick Klauschen,
  Klaus-Robert M{\"u}ller, and Wojciech Samek.
\newblock On pixel-wise explanations for non-linear classifier decisions by
  layer-wise relevance propagation.
\newblock {\em PloS one}, 10(7):e0130140, 2015.

\bibitem{ribeiro2016why}
Marco~Tulio Ribeiro, Sameer Singh, and Carlos Guestrin.
\newblock "why should i trust you?": Explaining the predictions of any
  classifier, 2016.

\bibitem{NIPS2017_0060ef47}
Piotr Dabkowski and Yarin Gal.
\newblock Real time image saliency for black box classifiers.
\newblock In I.~Guyon, U.~V. Luxburg, S.~Bengio, H.~Wallach, R.~Fergus,
  S.~Vishwanathan, and R.~Garnett, editors, {\em Advances in Neural Information
  Processing Systems}, volume~30, pages 6967--6976. Curran Associates, Inc.,
  2017.

\bibitem{RevModPhys.73.663}
J.~D. Jackson and L.~B. Okun.
\newblock Historical roots of gauge invariance.
\newblock {\em Rev. Mod. Phys.}, 73:663--680, Sep 2001.

\bibitem{grathwohl2020classifier}
Will Grathwohl, Kuan-Chieh Wang, Jörn-Henrik Jacobsen, David Duvenaud,
  Mohammad Norouzi, and Kevin Swersky.
\newblock Your classifier is secretly an energy based model and you should
  treat it like one, 2020.

\bibitem{RevModPhys.54.235}
F.~Y. Wu.
\newblock The potts model.
\newblock {\em Rev. Mod. Phys.}, 54:235--268, Jan 1982.

\bibitem{PhysRev.106.620}
E.~T. Jaynes.
\newblock Information theory and statistical mechanics.
\newblock {\em Phys. Rev.}, 106:620--630, May 1957.

\bibitem{PhysRev.108.171}
E.~T. Jaynes.
\newblock Information theory and statistical mechanics. ii.
\newblock {\em Phys. Rev.}, 108:171--190, Oct 1957.

\bibitem{lecun-mnisthandwrittendigit-2010}
Yann LeCun and Corinna Cortes.
\newblock {MNIST} handwritten digit database.
\newblock 2010.

\bibitem{morcos2011direct}
Faruck Morcos, Andrea Pagnani, Bryan Lunt, Arianna Bertolino, Debora~S Marks,
  Chris Sander, Riccardo Zecchina, Jos{\'e}~N Onuchic, Terence Hwa, and Martin
  Weigt.
\newblock Direct-coupling analysis of residue coevolution captures native
  contacts across many protein families.
\newblock {\em Proceedings of the National Academy of Sciences},
  108(49):E1293--E1301, 2011.

\bibitem{coucke2016direct}
Alice Coucke, Guido Uguzzoni, Francesco Oteri, Simona Cocco, Remi Monasson, and
  Martin Weigt.
\newblock Direct coevolutionary couplings reflect biophysical residue
  interactions in proteins.
\newblock {\em The Journal of chemical physics}, 145(17):174102, 2016.

\bibitem{muscat2020filterdca}
Maureen Muscat, Giancarlo Croce, Edoardo Sarti, and Martin Weigt.
\newblock Filterdca: Interpretable supervised contact prediction using
  inter-domain coevolution.
\newblock {\em PLoS computational biology}, 16(10):e1007621, 2020.

\bibitem{barrat2020sparse}
Pierre Barrat-Charlaix, Anna~Paola Muntoni, Kai Shimagaki, Martin Weigt, and
  Francesco Zamponi.
\newblock Sparse generative modeling of protein-sequence families.
\newblock {\em arXiv preprint arXiv:2011.11259}, 2020.

\bibitem{jacquin2016benchmarking}
Hugo Jacquin, Amy Gilson, Eugene Shakhnovich, Simona Cocco, and R{\'e}mi
  Monasson.
\newblock Benchmarking inverse statistical approaches for protein structure and
  design with exactly solvable models.
\newblock {\em PLoS computational biology}, 12(5):e1004889, 2016.

\bibitem{muntoni2020aligning}
Anna~Paola Muntoni, Andrea Pagnani, Martin Weigt, and Francesco Zamponi.
\newblock Aligning biological sequences by exploiting residue conservation and
  coevolution.
\newblock {\em Physical Review E}, 102(6):062409, 2020.

\bibitem{senior2020improved}
Andrew~W Senior, Richard Evans, John Jumper, James Kirkpatrick, Laurent Sifre,
  Tim Green, Chongli Qin, Augustin {\v{Z}}{\'\i}dek, Alexander~WR Nelson, Alex
  Bridgland, et~al.
\newblock Improved protein structure prediction using potentials from deep
  learning.
\newblock {\em Nature}, 577(7792):706--710, 2020.

\bibitem{yang2020improved}
Jianyi Yang, Ivan Anishchenko, Hahnbeom Park, Zhenling Peng, Sergey
  Ovchinnikov, and David Baker.
\newblock Improved protein structure prediction using predicted interresidue
  orientations.
\newblock {\em Proceedings of the National Academy of Sciences},
  117(3):1496--1503, 2020.

\bibitem{PhysRevE.87.012707}
Magnus Ekeberg, Cecilia L\"ovkvist, Yueheng Lan, Martin Weigt, and Erik Aurell.
\newblock Improved contact prediction in proteins: Using pseudolikelihoods to
  infer potts models.
\newblock {\em Phys. Rev. E}, 87:012707, Jan 2013.

\bibitem{harvey1964unidentified}
JJ~Harvey.
\newblock An unidentified virus which causes the rapid production of tumours in
  mice.
\newblock {\em Nature}, 204(4963):1104--1105, 1964.

\bibitem{downward2003targeting}
Julian Downward.
\newblock Targeting ras signalling pathways in cancer therapy.
\newblock {\em Nature Reviews Cancer}, 3(1):11--22, 2003.

\bibitem{malumbres2003ras}
Marcos Malumbres and Mariano Barbacid.
\newblock Ras oncogenes: the first 30 years.
\newblock {\em Nature Reviews Cancer}, 3(6):459--465, 2003.

\bibitem{lu2016ras}
Shaoyong Lu, Hyunbum Jang, Serena Muratcioglu, Attila Gursoy, Ozlem Keskin,
  Ruth Nussinov, and Jian Zhang.
\newblock Ras conformational ensembles, allostery, and signaling.
\newblock {\em Chemical reviews}, 116(11):6607--6665, 2016.

\bibitem{tensorflow2015-whitepaper}
Mart\'{\i}n Abadi, Ashish Agarwal, Paul Barham, Eugene Brevdo, Zhifeng Chen,
  Craig Citro, Greg~S. Corrado, Andy Davis, Jeffrey Dean, Matthieu Devin,
  Sanjay Ghemawat, Ian Goodfellow, Andrew Harp, Geoffrey Irving, Michael Isard,
  Yangqing Jia, Rafal Jozefowicz, Lukasz Kaiser, Manjunath Kudlur, Josh
  Levenberg, Dandelion Man\'{e}, Rajat Monga, Sherry Moore, Derek Murray, Chris
  Olah, Mike Schuster, Jonathon Shlens, Benoit Steiner, Ilya Sutskever, Kunal
  Talwar, Paul Tucker, Vincent Vanhoucke, Vijay Vasudevan, Fernanda Vi\'{e}gas,
  Oriol Vinyals, Pete Warden, Martin Wattenberg, Martin Wicke, Yuan Yu, and
  Xiaoqiang Zheng.
\newblock {TensorFlow}: Large-scale machine learning on heterogeneous systems,
  2015.
\newblock Software available from tensorflow.org.

\bibitem{simard2003best}
Patrice~Y Simard, David Steinkraus, John~C Platt, et~al.
\newblock Best practices for convolutional neural networks applied to visual
  document analysis.
\newblock In {\em Icdar}, volume~3, 2003.

\bibitem{bateman2004pfam}
Alex Bateman, Lachlan Coin, Richard Durbin, Robert~D Finn, Volker Hollich, Sam
  Griffiths-Jones, Ajay Khanna, Mhairi Marshall, Simon Moxon, Erik~LL
  Sonnhammer, et~al.
\newblock The pfam protein families database.
\newblock {\em Nucleic acids research}, 32(suppl\_1):D138--D141, 2004.

\bibitem{malinverni2019coevolutionary}
Duccio Malinverni and Alessandro Barducci.
\newblock Coevolutionary analysis of protein sequences for molecular modeling.
\newblock In {\em Biomolecular Simulations}, pages 379--397. Springer, 2019.

\end{thebibliography}

\section{Supplementary Materials}

\subsection{Explicit expansion for networks with one hidden layer}
We will denote with the superscript $[l]$ in square brackets the weights belonging to layer $l$ of a neural network.
The output $y$ of a classifier with a single hidden layer can be written as 
\begin{equation}\label{Eq:onelayer}
y=softmax\left(\W2\cdot f\left(\W1\cdot x+\B1\right)+\B2\right),
\end{equation}
where $W$ and $b$ are the weights and the biases respectively, and $f$ is a $C^\infty$ function used as activation function after the hidden layer.
The input vector $x$ is the concatenation of one-hot encodings for every input features. 
In order to simplify the notation, we will here label every entry of the input vector with a single subscript, e.g. $x_j$, instead of using $x_i^\alpha$ to represent category $\alpha$ of feature $i$.
We are interested in expanding in Taylor's series the argument to the softmax function, i.e. the Hamiltonians of the problem. 
The goal is to rewrite Equation \ref{Eq:onelayer} as
\begin{equation}\label{Eq:onelayerexp}
y=softmax\left( H_{0}+\sum_j J_{j}\;x_j + \sum_{j,k\neq j}J_{jk}\;x_j x_k+\cdots \right).
\end{equation}

We start by expanding the function $f$:
\begin{equation}\label{Eq:Taylor_f}
f_i\left(\W1\cdot x+\B1\right)=\sum_{n=0}^\infty \frac{1}{n!}f^{(n)}(0)\sum_{m=0}^n\binom{n}{m}\left(\sum_j\W1_{i,j}\;x_j\right)^m\;\left(\B1_{i}\right)^{(n-m)}
\end{equation}
and collecting the terms in increasing order of $x$.

We start by noticing that for every value of $n$ in Equation \ref{Eq:Taylor_f}, the only term that do not depend on $x$ is $m=0$, i.e.
\begin{equation}
\sum_{n=0}^\infty \frac{1}{n!}f^{(n)}(0)\left(\B1_{i}\right)^{n}=f(\B1_i).
\end{equation}

Also, by using the idempotency of $x$, it is easy to show that the linear term in $x$ correspond to
\begin{equation}
\sum_j\sum_{n=0}^\infty \frac{1}{n!}f^{(n)}(0)\sum_{m=1}^n\binom{n}{m}\left(\W1_{i,j}\right)^m\;\left(\B1_{i}\right)^{(n-m)}\;x_j,
\end{equation}
which can be rewritten as
\begin{equation}
\sum_j \left[f(\W1_{i,j}+\B1_{i})-f(\B1_i)\right]\;x_j.
\end{equation}
The quadratic order if given by
\begin{equation}
\sum_j\sum_k\sum_{n=2}^\infty \frac{1}{n!}f^{(n)}(0)\sum_{m=2}^{n-1}\binom{n}{m}\sum_{p=1}^{m-1}\binom{m}{p}\left(\W1_{i,j}\right)^p\left(\W1_{i,k}\right)^{m-p}\;\left(\B1_{i}\right)^{(n-m)}\;x_j\;x_k,
\end{equation}
which can be simplified as
\begin{equation}
\sum_j\sum_k\left[ f(\W1_{i,j}+\W1_{i,k}+\B1_{i})-f(\W1_{i,j}+\B1_{i})-f(\W1_{i,k}+\B1_{i})+f(\B1_{i}) \right]\;x_j\;x_k.
\end{equation}

We can finally compute the coefficients of the expansion \ref{Eq:onelayerexp} as:
\begin{equation}
\begin{split}
H_{0,\ell} &=\sum_i \W2_{\ell,i}\; f(\B1_i)+\B2_\ell \\
J_{j,\ell} &=\sum_i \W2_{\ell,i}\; \left[f(\W1_{i,j}+\B1_{i})-f(\B1_i)\right] \\
J_{jk,\ell}&=\sum_i \W2_{\ell,i}\; \left[ f(\W1_{i,j}+\W1_{i,k}+\B1_{i})-f(\W1_{i,j}+\B1_{i})-f(\W1_{i,k}+\B1_{i})+f(\B1_{i}) \right]
\end{split}
\end{equation}

\newpage

\subsection{2 bits MNIST dataset}
\begin{figure}[h!]
\begin{center}
\includegraphics[width=0.9\textwidth]{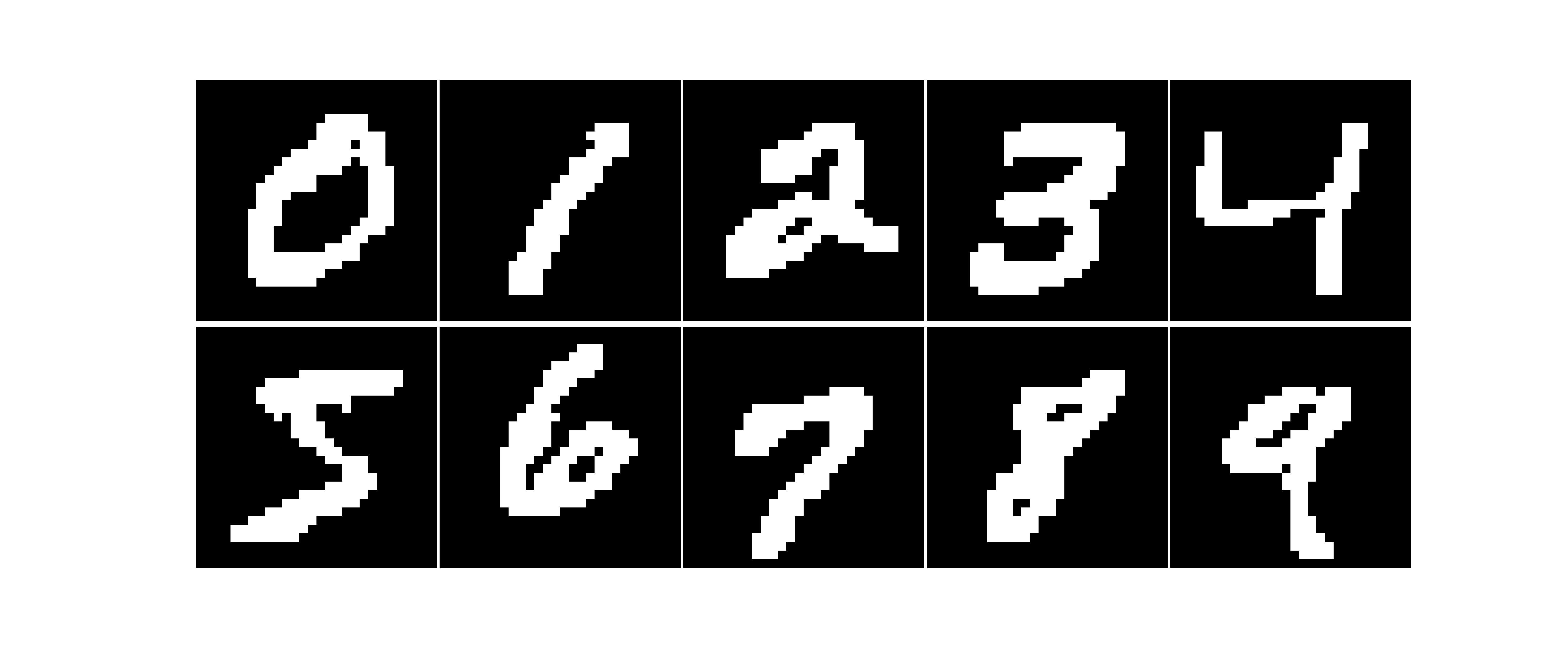}
\caption{Example digits in our 2bit dataset}
\label{fig:2bitdataset}
\end{center}
\end{figure}

\begin{figure}[h!]
\begin{center}
\includegraphics[width=0.9\textwidth]{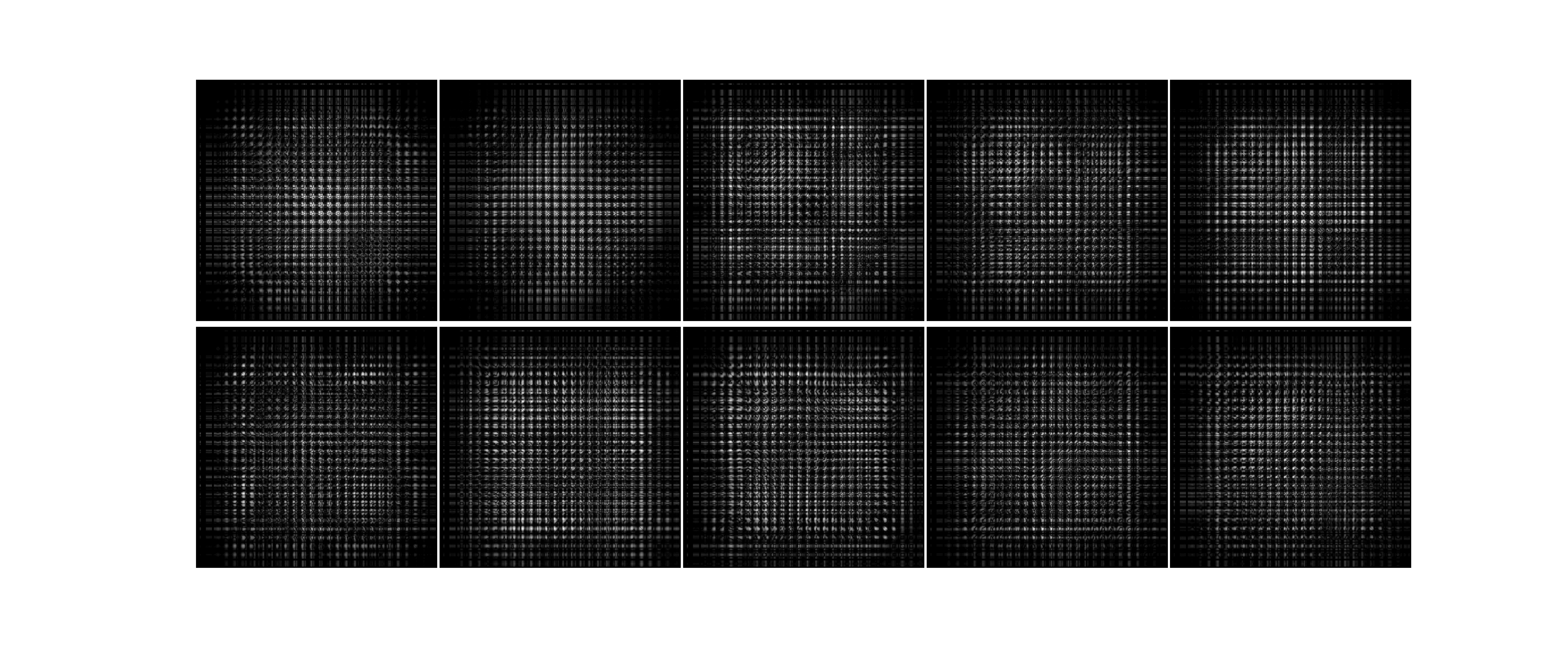}
\caption{Alternative representation of the second order scores. Each pixel directly represents one coefficient of the expansion. Black is zero. Each row (column) $28*n+m$ comprises all coefficients relative to pixel $(n,m)$ in the original image. Each image is 784x784 pixels. }
\label{fig:2bitdataset2dim}
\end{center}
\end{figure}

\newpage
\subsection{4 bits MNIST dataset}

\begin{figure}[h!]
\begin{center}
\includegraphics[width=0.9\textwidth]{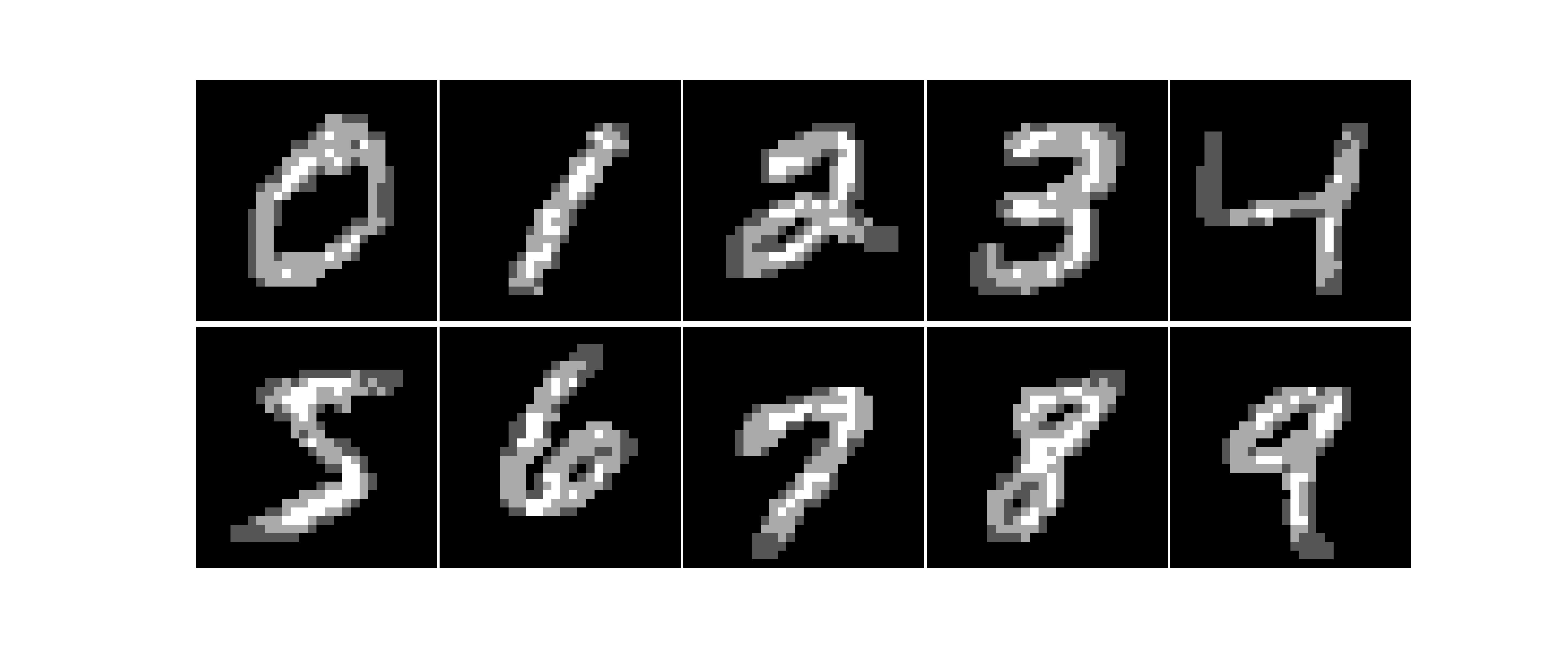}
\caption{Example digits in our 4bit dataset}
\label{fig:4bitdataset}
\end{center}
\end{figure}

\begin{figure}[h!]
\begin{center}
\includegraphics[width=0.9\textwidth]{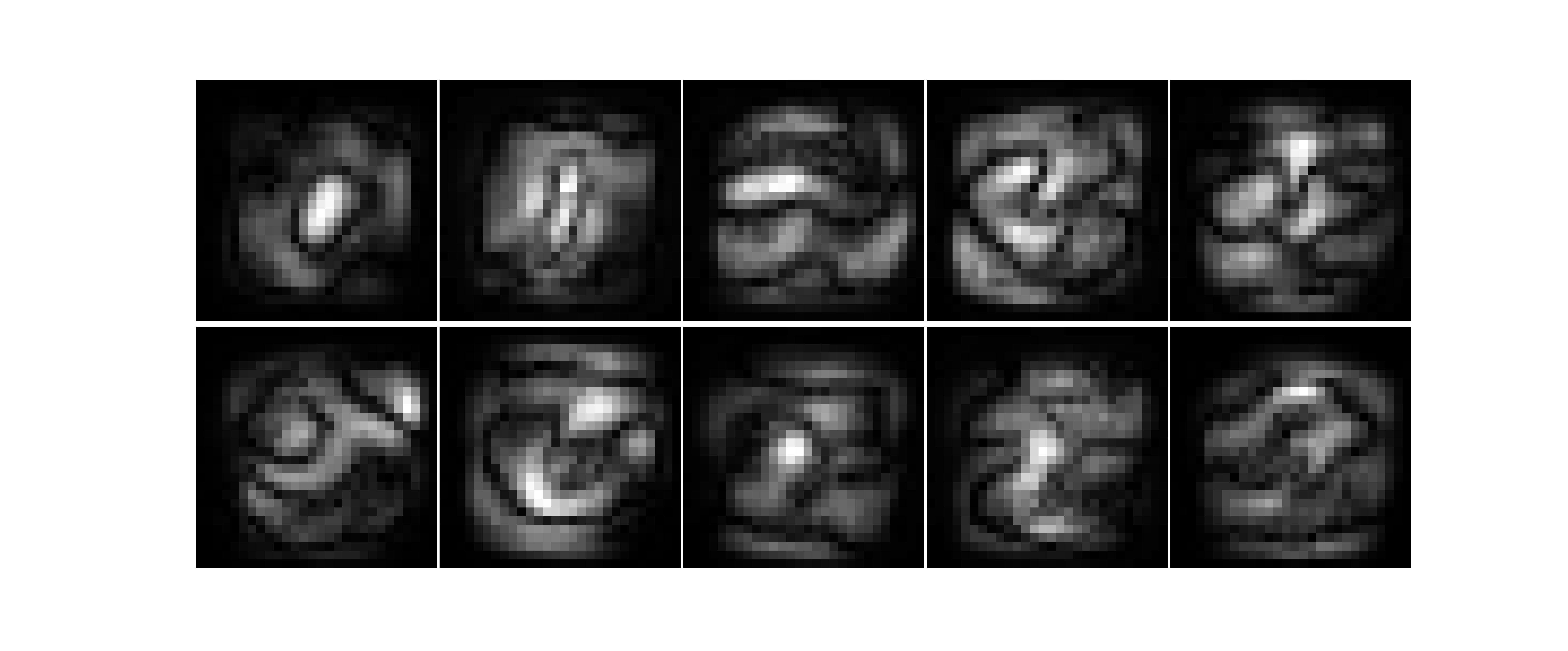}
\caption{Scores from first order expansion of the Energy learned from our 4-bits MNIST dataset. Black corresponds to zero. Features that resembles the corresponding digits are clearly visible}
\label{fig:4bits_score1}
\end{center}
\end{figure}

\begin{figure}[h!]
\begin{center}
\includegraphics[width=0.9\textwidth]{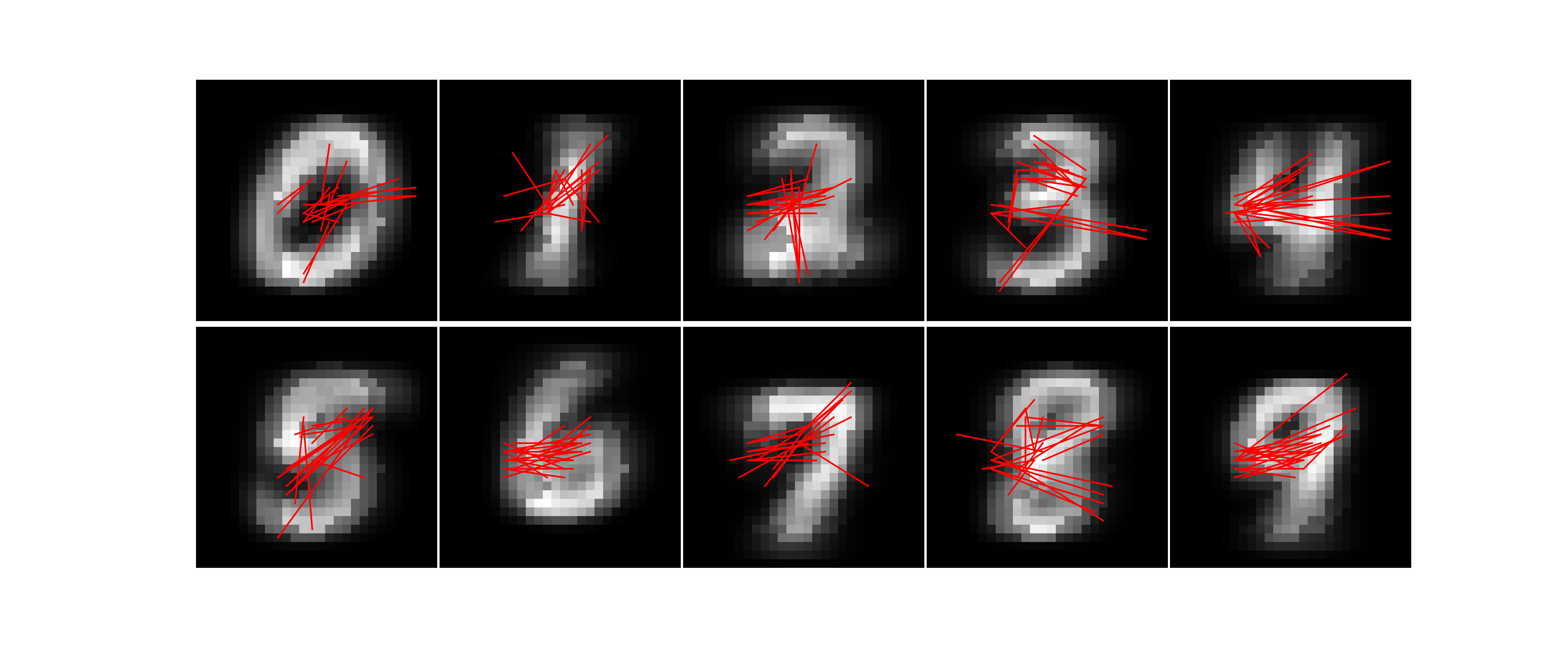}
\caption{Representation of the second order scores of the network trained with our 4-bits MNIST dataset. Red lines connect the 5 highest scored non adjacent pixel pairs.}
\label{fig:4bitdataset2dimlines}
\end{center}
\end{figure}

\begin{figure}[h!]
\begin{center}
\includegraphics[width=0.9\textwidth]{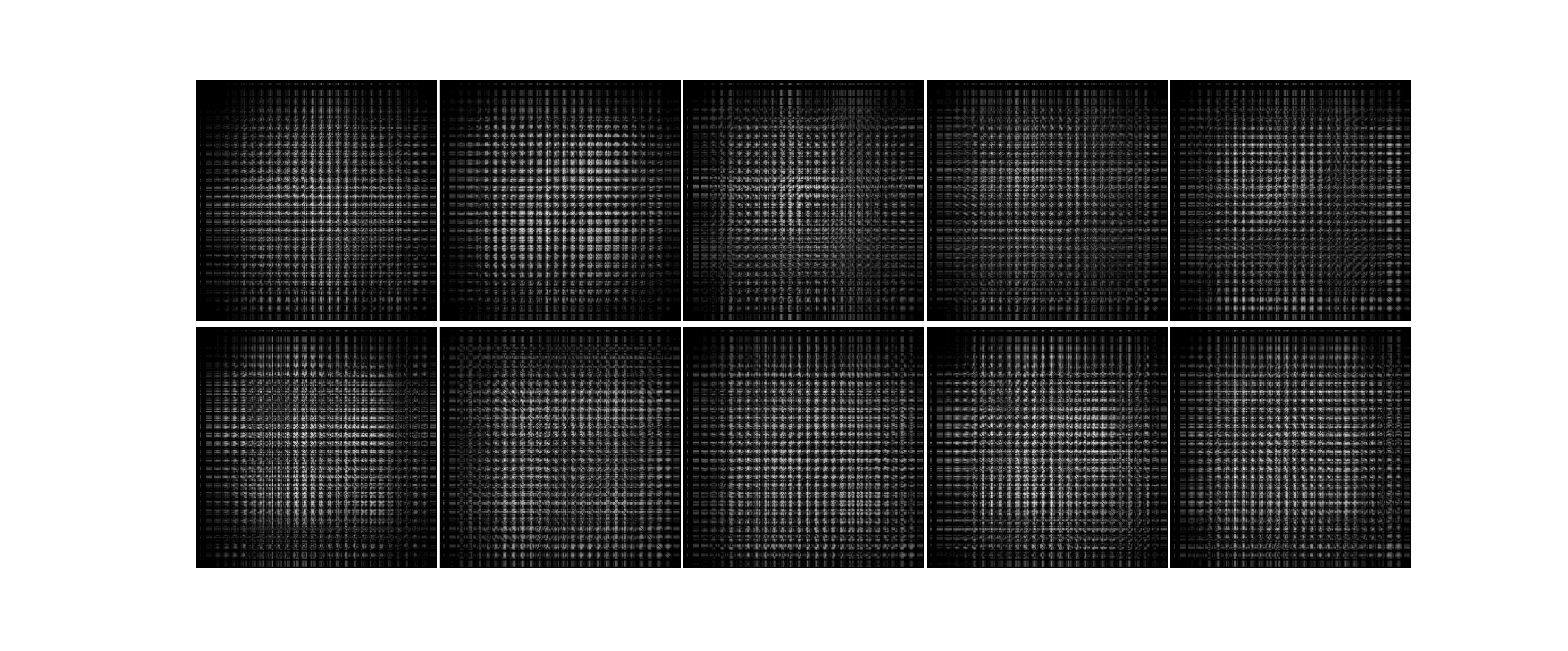}
\caption{Alternative representation of the second order scores of the network trained with our 4-bits MNIST dataset. Each pixel directly represents one coefficient of the expansion. Black is zero. Each row (column) $28*n+m$ comprises all coefficients relative to pixel $(n,m)$ in the original image. Each image is 784x784 pixels. }
\label{fig:4bitdataset2dim}
\end{center}
\end{figure}

\clearpage

\subsection{Direct coupling analysis of RAS family}

\end{document}